\documentclass[runningheads]{llncs}
\usepackage{lipsum}
\usepackage{multirow}
\usepackage{pgfplots}

\usepackage{eccv}

\usepackage{eccvabbrv}

\usepackage{graphicx}
\usepackage{booktabs}

\usepackage[accsupp]{axessibility}  

\usepackage[pagebackref,breaklinks,colorlinks]{hyperref}

\newif\ifdoubleblind
\doubleblindfalse

\usepackage{orcidlink}

\begin{document}

\title{Deepfake Detection without Deepfakes:\\ Generalization via Synthetic\\ Frequency Patterns Injection}

\titlerunning{Deepfake Detection without Deepfakes}

\author{Davide Alessandro Coccomini\inst{1}\orcidlink{0000-0002-0755-6154} \and
Roberto Caldelli\inst{2,3}\orcidlink{0000-0003-3471-1196} \and
Claudio Gennaro\inst{1}\orcidlink{0000-0002-3715-149X} \and\\
Giuseppe Fiameni\inst{4}\orcidlink{0000-0001-8687-6609} \and
Giuseppe Amato\inst{1}\orcidlink{0000-0003-0171-4315} \and
Fabrizio Falchi\inst{1}\orcidlink{0000-0001-6258-5313}
}

\authorrunning{D.A.~Coccomini et al.}

\institute{ISTI-CNR, Pisa, Italy\\
\and
CNIT, Florence, Italy
\and
Universitas Mercatorum, Rome, Italy
\and
NVIDIA AI Technology Center, Italy
\email{\{davidealessandro.coccomini,claudio.gennaro,\\giuseppe.amato,fabrizio.falchi\}@isti.cnr.it}\\
\email{roberto.caldelli@cnit.it}\\
\email{gfiameni@nvidia.com}
}
\maketitle

\begin{abstract}
Deepfake detectors are typically trained on large sets of pristine and generated images, resulting in limited generalization capacity; they excel at identifying deepfakes created through methods encountered during training but struggle with those generated by unknown techniques.
This paper introduces a learning approach aimed at significantly enhancing the generalization capabilities of deepfake detectors. Our method takes inspiration from the unique ``fingerprints'' that image generation processes consistently introduce into the frequency domain. These fingerprints manifest as structured and distinctly recognizable frequency patterns. We propose to train detectors using only pristine images injecting in part of them crafted frequency patterns, simulating the effects of various deepfake generation techniques without being specific to any. These synthetic patterns are based on generic shapes, grids, or auras.
We evaluated our approach using diverse architectures across 25 different generation methods. 
The models trained with our approach were able to perform state-of-the-art deepfake detection, demonstrating also superior generalization capabilities in comparison with previous methods. Indeed, they are untied to any specific generation technique and can effectively identify deepfakes regardless of how they were made.

To use the proposed approach and reproduce the presented experiments, the code is available at:
\ifdoubleblind
[Hidden for double-blind review]
\else
\url{https://github.com/davide-coccomini/Deepfake-Detection-without-Deepfakes-Generalization-via-Synthetic-Frequency-Patterns-Injection}
\fi
\end{abstract}


\section{Introduction}
\label{sec:intro}
Image synthesis techniques are more accessible and powerful than ever, making distinguishing a generated image from a pristine one often quite a complex task, easily risking being misled and becoming a victim of misinformation. These contents, also known as deepfakes, are therefore a serious risk to society. Several deepfakes detectors have been developed to face this problem and they have proven to be extremely effective in discriminating between reality and fiction, especially in the case of fully synthetic images\cite{coccominidiffusion,amoroso,Dogoulis_2023,uninadms}. However, there remains a serious open problem in the field, that of generalization. It is in fact, quite common that deepfake detectors trained to distinguish pristine and fake images are totally ineffective when they have to classify images generated by methods unused in their training set construction.
It, therefore, appears that rather than learning the generic concept of deepfake, the detectors learn to recognize some sort of trace or artifact specific to the model used to generate the training images\cite{crossforgery, crossforgery2,reverse, coccominidiffusion}. This translates into their total inadequacy in facing the daily emergence of new generation methods, finding themselves unprepared and potentially needing continuous re-training.

A possible solution may reside in the exploitation of the findings of several studies which showed that synthetic image generators, whether based on Generative Adversarial Networks (GANs)\cite{gan} or Diffusion Models (DMs)\cite{ddpm}, tend to introduce specific fingerprints in the frequency domain\cite{uninadms,uninagans,reverse}. It is, therefore, plausible to think that deepfake detectors are simply learning to recognize these fingerprints, and when they see a new one, just classify the image as pristine as it does not correspond to any of the fingerprints learned in the training phase.

To highlight the fingerprints' peculiarities and propose a way to exploit them, we first present an in-depth analysis of the traces introduced by various image-generation models. From our observations, these fingerprints, although differing from model to model, share the characteristic of being structured patterns (geometric shapes, grids, etc.). We, therefore, propose a novel deepfake detection technique consisting of training a model to recognize between pristine images, taken from MSCOCO\cite{mscoco}, and images to which a synthetic structured pattern has been injected in the frequency domain. These patterns, inspired by geometric shapes, grids, or auras and unrelated to any particular generation method, are thought to stimulate the model to search for artifacts in the frequency domain disincentivizing the memorization of specific fingerprints. The models trained with this strategy demonstrated state-of-the-art capability in the detection of fake images regardless of the generator used to obtain the images.

\section{Related Works}

\subsection{Deepfake Generation}
The creation of synthetic images can today be performed through multiple techniques, from methods based on Generative Adversarial Networks \cite{stylegan,stylegan2,biggan,bicyclegan,gan} to more recent ones based on Diffusion Models \cite{ldm,glide,ddim,sdxl}. In the first case, images are created by a generator model that is posed in an adversarial context against a discriminator model that evaluates the image synthesized from time to time by the generator and tries to figure out whether it is fake or not. In doing so, the generator will try to create gradually more credible images until it has succeeded in fooling the discriminator. On the other hand, Diffusion Models are a type of model that can transform a starting distribution into more complex structures depending on the task of training. In both cases, the image generation process tends to introduce visible artifacts in the spatial domain and, most importantly, even in the more advanced models, a visualizable fingerprint in the frequency domain specific to the generator used\cite{uninagans,uninadms,reverse}.

\subsection{Deepfake Detection}
Identifying deepfakes is a task rather addressed in the literature with studies proposing models able to manage images by analyzing only the spatial aspect such as \cite{coccominicombining,wodajo2021deepfake} or models that can also consider the temporal aspect in the case of videos such as \cite{mintime,ftcn,caldelli}. These models tend to focus on detecting artifacts and inconsistencies introduced by manipulation techniques. Recently, with the advent of diffusion models, many deepfake detectors have also begun to emerge focusing on the more specific topic of synthetic image detection, such as \cite{amoroso, coccominidiffusion,Dogoulis_2023, uninadms}. These methods are mainly trained in a binary classification setup to distinguish between pristine and fake images or videos. This detection may be improved using approaches such as the one proposed in \cite{Giudice_2021} who pointed out that searching for artifacts in the frequency domain exploiting techniques such as the Discrete Cosine Transform (DCT), can be quite effective in deepfake detection.
Unfortunately, as highlighted in \cite{crossforgery, crossforgery2, jimaging8100263}, deepfake detectors tend to learn to recognize images generated by methods used for constructing the training set almost exclusively and fail to generalize the concept of deepfake.
In this paper, we then try to face the issue of generalization in synthetic image detection by proposing a different approach in model training, stimulating the understanding of the more general concept of deepfake.

\section{Method}
The proposed training method for deepfake detection models involves injecting synthetic frequency patterns into pristine images. The models are trained in a binary classification framework to discriminate between pristine images and those with injected patterns (labelled as fake). This approach serves as a prototype for deepfake detection, aiming to enhance the model's generalization capabilities.

\subsection{Fingerprint Extraction}
\label{sec:fingerprint_extraction}
\begin{figure}[t]  
  \centering
  \includegraphics[width=\linewidth]{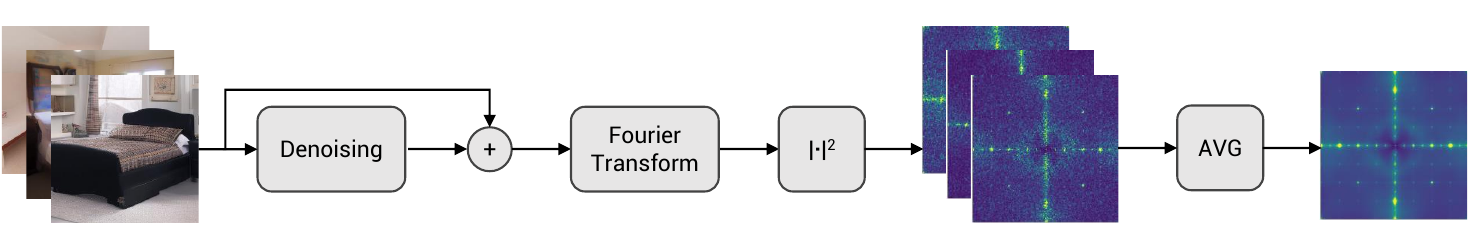} 
  \caption{The fingerprints extraction process presented in \cite{uninagans} is shown in the figure. Considering a set of images obtained from the same generator, they undergo a denoising process and then are transformed to the frequency domain by Fourier Transform. The square of the Fourier Transform is averaged to obtain a unique generator's fingerprint.}
  \label{fig:fingerprint_extraction}
\end{figure}
To observe the fingerprints introduced in the synthetic images by image generators and thus inspire the synthetic pattern generation on which our method is based, we replicated the procedure from \cite{uninagans} that we summarize in Figure \ref{fig:fingerprint_extraction}. 

The approach proposed by the authors analyzes a set of images created by a generator to extract its fingerprint. On each image, we perform a denoising and normalization step to sharpen the pattern traces. Their Power Spectrum is then extracted to highlight the frequency energy peaks. Finally, we calculate the average of all power spectrums obtaining the generator's fingerprint.
Using this procedure, it is then possible to extract and observe its peculiarities.

\begin{figure}[t]
  \centering
  \begin{subfigure}[b]{\textwidth}
    \centering
    \includegraphics[width=\textwidth]{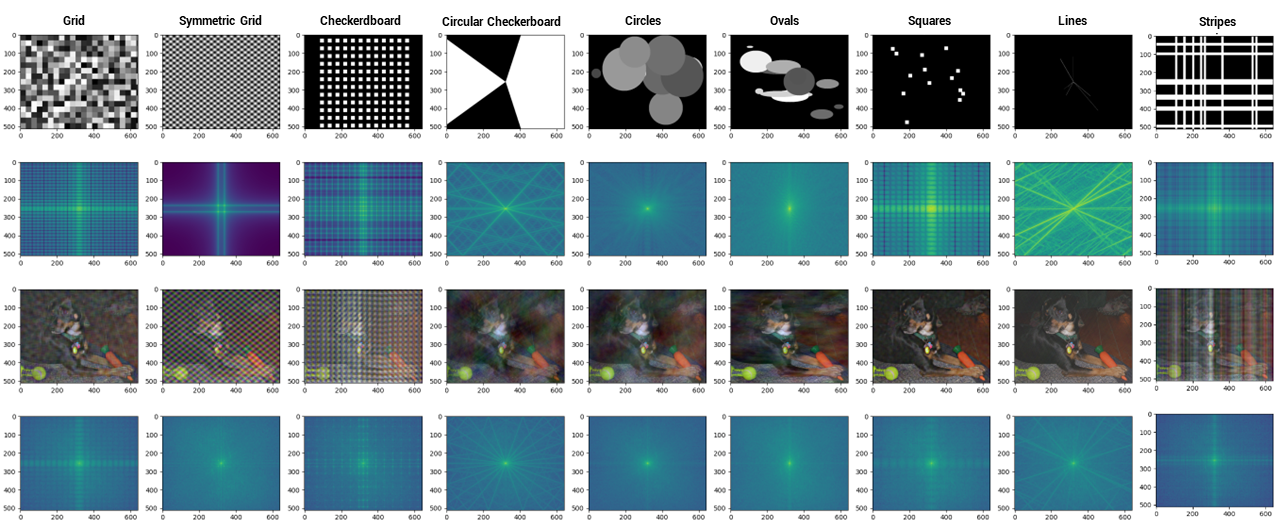}
    \caption{Geometric patterns}
    \label{fig:geometric}
  \end{subfigure}

  \vspace{1em}
\begin{subfigure}[b]{0.7\textwidth}
    \centering
    \includegraphics[width=\textwidth]{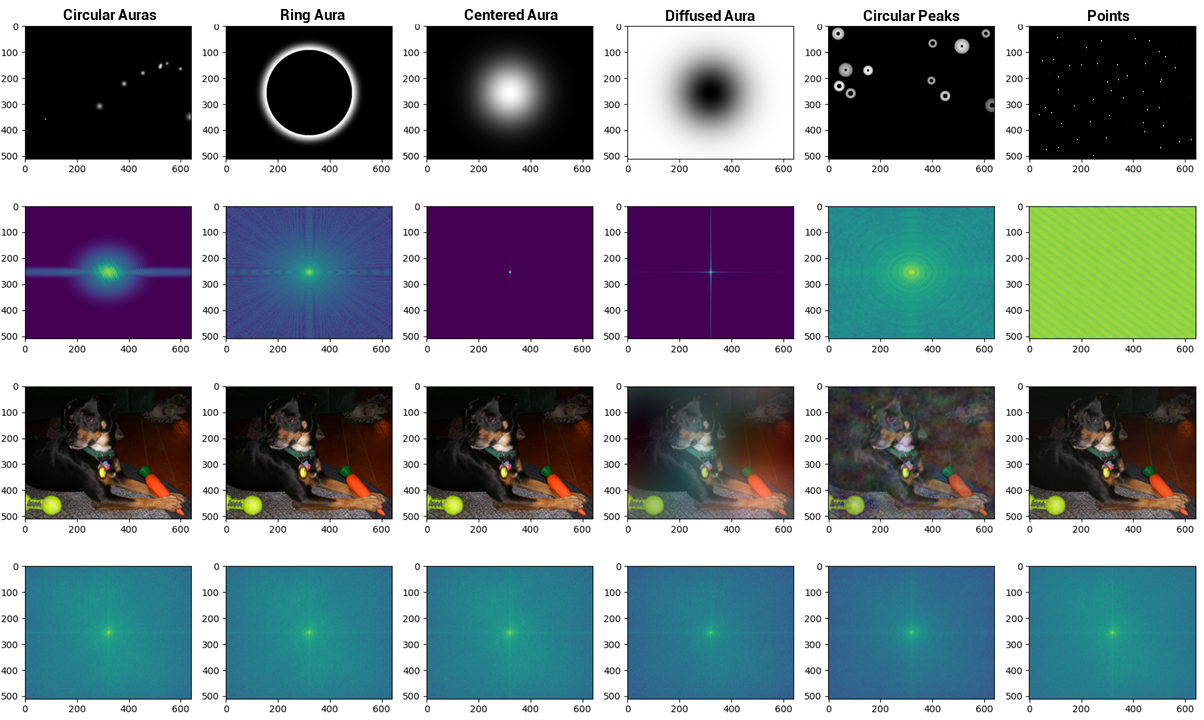}
    \caption{Aura patterns}
    \label{fig:auras}
\end{subfigure}\hfill%
\begin{subfigure}[b]{0.26\textwidth}
    \centering
    \includegraphics[width=\textwidth]{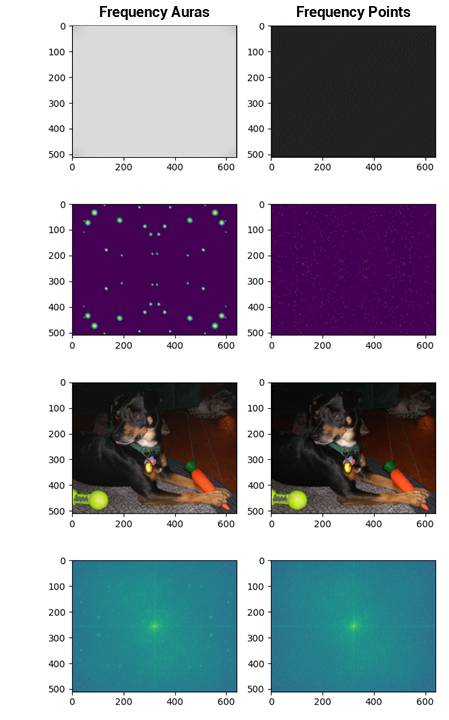}
    \caption{Spike patterns}
    \label{fig:fourier}
\end{subfigure}

  \caption{Overview of synthetically generated patterns. For each category of pattern, the first row indicates the pattern generated in the spatial domain, the second is the magnitude of its Fourier Transform, the third row shows the pristine image on which the pattern has been applied, and the last row shows its Fourier Transform.}
  \label{fig:patterns_overview}
\end{figure}

\subsection{Patterns Generation}
\label{sec:pattern_generation}
Taking as inspiration the common features of fingerprints observed from our analysis and previous works in this area \cite{uninagans,uninadms,reverse}, we created a pattern generator capable of synthesizing different kinds of patterns. We show in Figure \ref{fig:patterns_overview} all the designed patterns that can be applied to a sample image and its representation in the frequency domain. The patterns are different at each generation thanks to the randomization of the characteristics specific to each of them, while always remaining structured. More in detail, they can be grouped into three categories:
\begin{itemize}
    \item \emph{Geometric}: These are geometric patterns generated in the spatial domain, such as grids, checkerboards, circles, ovals or stripes created in random numbers, intensities and positions with each generation.
    \item \emph{Auras}: These are peaks and auras generated in the spatial domain that can be randomly placed in the image in more or less variable numbers and of varied shapes such as rings or dots.
    \item \emph{Frequency Spikes}: These are patterns that in the spatial domain have no relevant spatial connotation but in the frequency domain generate symmetrical intensity spikes scattered throughout the image.
\end{itemize}

We empirically observed that these patterns are not related to any specific deepfake's fingerprint but are all structured and pretty rare to find in pristine images. This lack of correlation with actually generated images is the peculiarity that allows us to boost models' generalization. Our method can be applied using a virtually unlimited number of patterns, not having to use only those proposed.

\begin{figure}[t]  
  \centering
  \includegraphics[width=\linewidth]{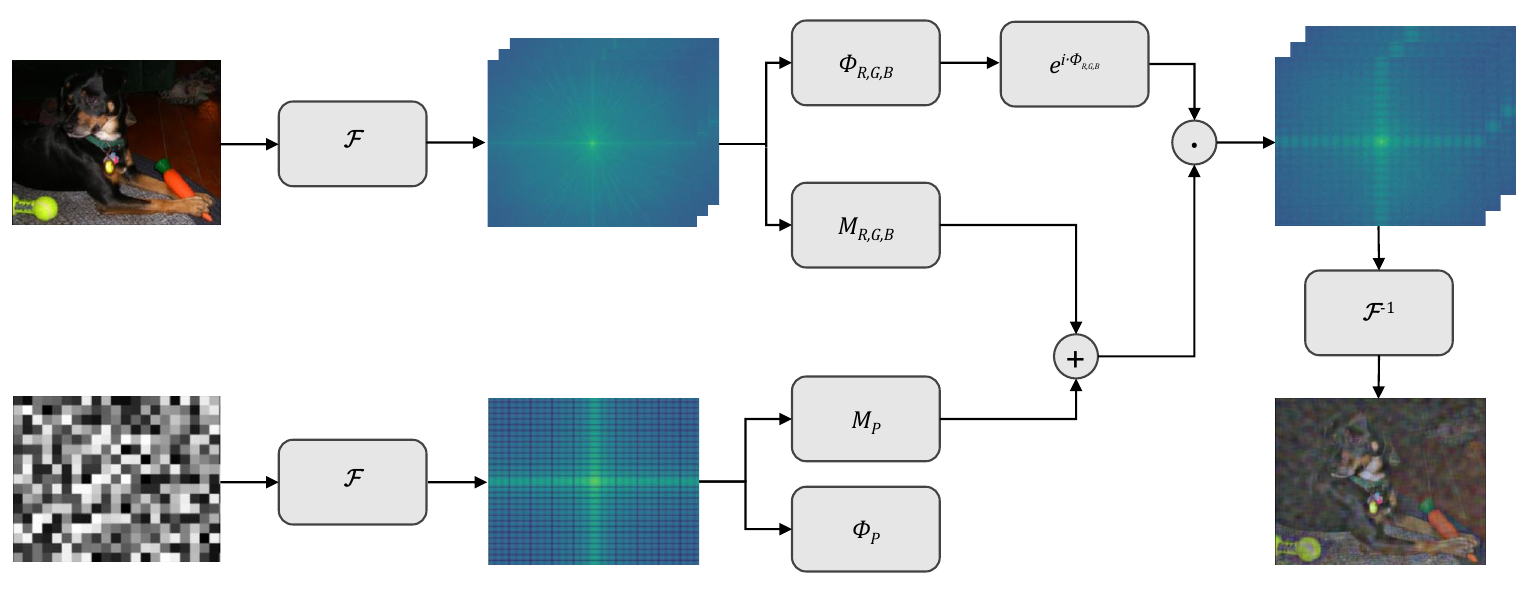} 
  \caption{The process of applying a pattern to an image is shown in the figure. The Fourier Transforms of both are first calculated. From them, we derive the magnitude and phase of the image and pattern. The pattern's magnitude is added to the magnitude of each image channel. Combining these components, we obtain the image with the pattern applied.}
  \label{fig:pattern_application}
\end{figure}

\subsection{Patterns Injection }
\label{sec:pattern_application}

After generating a synthetic pattern, it is necessary to effectively inject it into the image to turn it into a fake sample. In Figure \ref{fig:pattern_application} we visually represent the procedure of applying the generated pattern to an image. 
Formally, considering a pristine image $I$ we calculate the Fourier Transform $\mathcal{F}$ of each of its channels. We then generate a pattern $P$ and calculate its Fourier Transform too, as follows:

\begin{equation}
 F_{I_c} = \mathcal{F}(I_c) \hspace{1em} \forall c  \in \{R, G, B\}, \hspace{4em} F_p= \mathcal{F}(P)
\end{equation}
The pattern is generated in the spatial domain but we work in the frequency domain as this step allows the pattern to become less artifactual while still maintaining its characteristics and allowing us to work with separated magnitude and phase.
We then calculate the magnitude $M_P$ and phase $\phi_P$ of the pattern $P$  and the magnitude $M_{I_c}$ and phase $\phi_{I_c}$ of each image's channel: 
\begin{equation}
\begin{array}{c@{}l}
M_P = \|F_P\|, \quad \phi_P = \text{arg}(F_P)\\\\
M_{I_c} = \|F_{I_c}\| \hspace{1em}, \quad \phi_{I_c} = \text{arg}(F_{I_c}) \quad \forall c \in \{R, G, B\}
\end{array}
\end{equation}

Since we want to preserve the original image representation as much as possible, while still obtaining a frequency manipulation, in combining all the information obtained in the previous steps, we ignore the pattern phase to give more weight to the original image phase, this being the most informative part. Also, given that the footprint extraction conducted by Corvi et al. \cite{uninadms} is based on the energy density of the signal, our synthetic patterns exclusively perturb the magnitude component of the signal in the frequency domain.

Therefore we combine each image channel $I_c$ and the pattern $P$ in the frequency domain obtaining the Combined Fourier Transforms $F_{comb_c}$ as follows:

\begin{equation}
F_{comb_c} = M_{comb_c} \cdot e^{i \cdot \phi_{I_c}} \quad \forall c \in {R,G,B}
\end{equation}

where $M_{comb_c}$ is obtained by adding the pattern's magnitude to that of each image channel, as follows:

\begin{equation}
M_{comb_c} = M_P + M_{I_c} \quad \forall c \in {R,G,B}
\end{equation}

Finally, we stack together the Combined Fourier Transforms of all the channels and calculate its inverse Fourier Transform to return in the spatial domain and obtain the image with the applied pattern $I_P$:

\begin{equation}
I_P = [\mathcal{F}^{-1}(F_{comb_c}) | c \in \{R,G,B\}]
\end{equation}
This process has a visual impact on the image in the spatial domain by generating artifacts and in the frequency domain where a combined spectrum between the pattern and the initial image is obtained. The resulting image will then have within it the frequency domain structures of various kinds arising from the generated pattern.

\begin{figure}[t]  
  \centering
  \includegraphics[width=\linewidth]{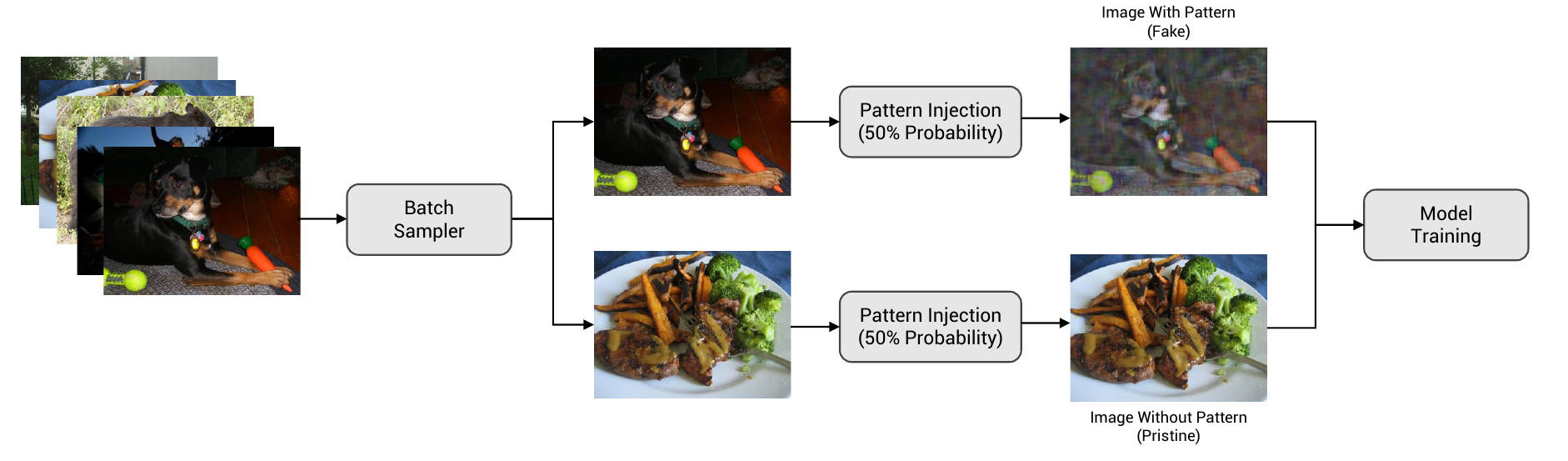} 
  \caption{The proposed training procedure is shown in the figure. During the training data loading phase, each image in the considered batch randomly undergoes pattern injection. If this occurs, the image is considered fake, otherwise it remains pristine.}
  \label{fig:training_procedure}
\end{figure}
\subsection{Training procedure}
Traditionally, deepfake detectors are trained on large datasets containing both pristine and fake images generated through one or more methods. 
In the proposed setup we define the classification problem as differentiating between pristine images and those that have been altered through frequency injection with one of the generated patterns, as shown in Section \ref{sec:pattern_generation}.

Therefore, considering a training set of only pristine images, during the model's training, whenever a sample is loaded for batch construction, it may randomly undergo the injection of a synthetic pattern through the process presented in Section \ref{sec:pattern_application}. If this happens, we assign the label "fake" to the transformed sample, otherwise, the image remains considered "pristine". Statistically, then in each batch, half of the samples will be pristine while the other half will have been injected with a synthetic pattern. In Figure \ref{fig:training_procedure} we summarize this procedure.
The model is then stimulated to learn to recognize the generic concept of structured frequency patterns within images without being in any way connected to a specific image generation method. 

\section{Experiments}
\subsection{Experimental Setup}
To carry out our experiments, we trained two separate deepfake detectors, based on Resnet50\cite{resnet} or on Swin-Base\cite{swin}, both of them are pretrained on Imagenet\cite{imagenet}. The models were trained 
on four NVIDIA GPU A100s with a batch size of 512, a learning rate of 0.01 decaying through a cosine scheduler to a minimum of 0.001 and using Binary Cross Entropy loss. Images are randomly augmented through geometric transformations such as rotation or reflection, light transformations such as contrast and color alteration, and others like the use of DCT or the application of noise. As explained in more detail in Section \ref{sec:classification}, the trained models demonstrated adequate performance exclusively after the first epoch thereafter tending to specialize in the training task which is different from the validation and test task. Therefore, the models reported in the experiments performed only one epoch of training on the dataset.

%


\begin{figure}[t]
  \begin{subfigure}[c]{.65\linewidth}
    \centering
    \includegraphics[width=\linewidth]{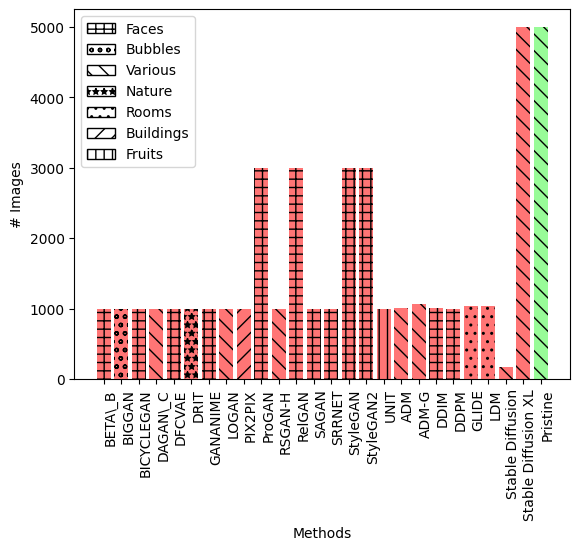}%
    \caption
      {%
        Number of images per generator.%
        \label{fig:datasets1}%
      }%
  \end{subfigure}\hfill
  \begin{tabular}[c]{@{}c@{}}
    \begin{subfigure}[c]{0.25\linewidth}
      \centering
      \includegraphics[width=\linewidth]{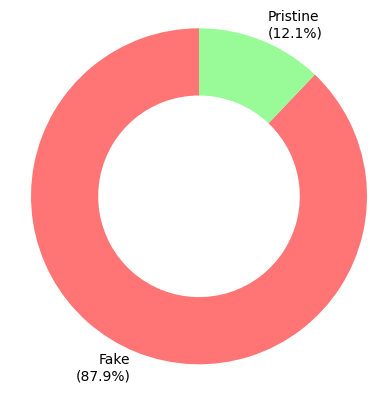}%
      \caption
        {%
          Percentage of fake and pristine images%
          \label{fig:datasets2}%
        }%
    \end{subfigure}\\
    \noalign{\bigskip}%
    \begin{subfigure}[c]{.3\linewidth}
      \centering
      \includegraphics[width=\linewidth,page=2]{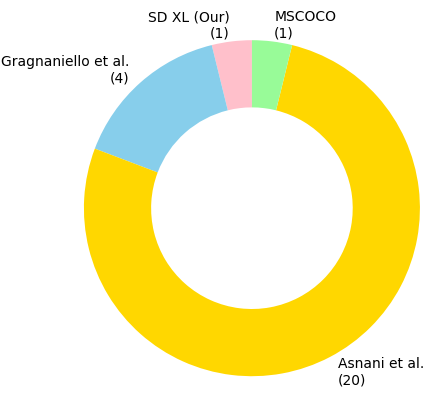}%
      \caption
        {%
          Distribution of the 25 considered generation methods.
          \label{fig:datasets3}%
        }%
    \end{subfigure}
  \end{tabular}
    {%
      
  \caption{Statistics of the constructed test set about the number of images per generator method with their subjects (a), percentage of pristine and fake images (b) and number of generators per datasets' authors (c).}%
      \label{fig:datasets}%
    }
\end{figure}
\subsection{Dataset}
\label{sec:dataset}
To perform the models' training, we used exclusively the around 118,000 pristine images from MSCOCO\cite{mscoco} training set. Since the pattern injection is randomly conducted at training time, half of the training images are untouched remaining considered pristine while the other half, namely the injected ones, compose the fake class.  The test set was obtained by combining the data presented in \cite{uninagans} and \cite{reverse} comprising images generated by 25 methods of both Generative Adversarial Networks and Diffusion Models. More in detail, we used the images obtained with StyleGAN (CelebAHQ)\cite{stylegan}, StyleGAN2\cite{stylegan2}, ProGAN\cite{progan} and RelGAN\cite{relgan} from \cite{uninagans} and images created with BETA\_B, BigGAN\cite{biggan}, BicycleGAN\cite{bicyclegan}, DAGAN\_C\cite{daganc}, DFCVAE\cite{dfcvae}, DRIT\cite{DRIT}, GANANIME\cite{gananime}, LOGAN\cite{logan}, PIX2PIX\cite{pix2pix}, SAGAN\cite{sagan}, SRRNET, RSGAN-H\cite{rsgan}, UNIT\cite{unit}, ADM\cite{adm}, ADM-G\cite{adm}, DDIM\cite{ddim}, DDPM\cite{ddpm}, GLIDE\cite{glide}, LDM\cite{ldm} and Stable Diffusion\cite{ldm} taken from the test set of \cite{reverse}. In addition, using the MSCOCO validation set captions, we generated 5,000 images using Stable Diffusion XL\cite{sdxl}. For the pristine class of the test set, we considered the 5,000 images from the MSCOCO validation set. 
Figure \ref{fig:datasets1} shows the number of images for each generator used in the construction of the test set and the kind of subject mainly depicted in the images generated by them (faces, buildings, etc.). The test set is then composed of more than 36,000 fake images and 5,000 pristine ones. In terms of proportion, as shown in Figure \ref{fig:datasets2}, the fake images cover almost 88\% of the dataset, thus making it unbalanced. Finally, in Figure \ref{fig:datasets3}, the contribution in terms of the number of image generation methods is shown for each author of the datasets considered.  
Therefore, the test set constructed covers a very wide range of methods and allows us to fully evaluate the generalization of our models by considering both GAN-based and Diffusion-based generators.

\subsection{Metrics}
To assess the effectiveness of the proposed method, we selected some metrics appropriate concerning the test set constructed and presented in Section \ref{sec:dataset}.
For each model considered, we reported the results obtained on each image generation method (or on pristine images) that compose the test set using the following metrics:
\begin{equation}
    Recall = \frac{TP}{TP+FN}, \quad\quad Specificity = \frac{TN}{TN+FP}
\end{equation}

where $TP$ is the number of true positive (fake) images, and $TN$ is the number of true negative (pristine) images. $FN$ and $FP$ represent the number of false negatives and false positives respectively. For a more fine-grained evaluation, we report these metrics on subsets composed of only fake or only pristine images. 
The threshold used to differentiate between the fake and pristine classes is consistently set at 0.5.
As the test set is unbalanced, to give a general idea of how the models behave on the overall dataset, we also visualized the ROC Curves and calculated the corresponding AUC scores. 



\begin{figure}[t]  
  \centering
  \includegraphics[width=\linewidth]{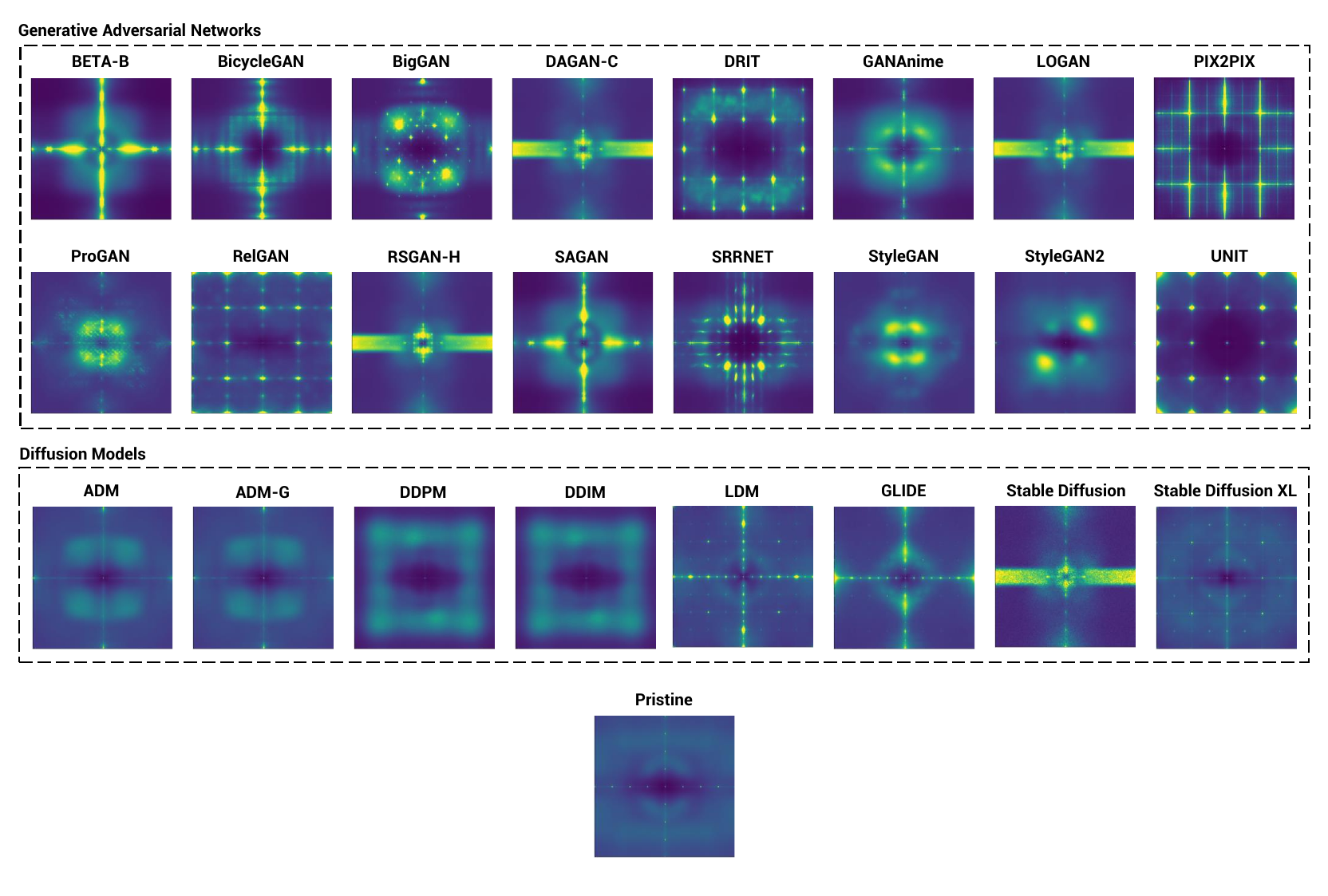} 
  \caption{Visualization of the fingerprints extracted through the process presented in \cite{uninagans} from the test set generators and pristine images.}
  \label{fig:fingerprints_overview}
\end{figure}

\subsection{Fingerprints Analysis}
\label{sec:fingerprint_analysis}
As a preliminary study, using the procedure explained in Section \ref{sec:fingerprint_extraction}, we extracted the fingerprints of all the methods composing the test set presented in Section \ref{sec:dataset}. As can be seen from Figure \ref{fig:fingerprints_overview}, the fingerprints of the various generators are very different from each other but tend to have specific, structured, and often grid-like shapes.
In particular, this shape is highly visible for the DRIT, PIX2PIX, RelGAN, SRRNET, UNIT, LDM and Stable Diffusion XL methods. In contrast, in the BETA\_B, BigGAN, GANAnime, ProGAN, SAGAN, StyleGAN, StyleGAN2, ADM, ADM-G, DDPM, DDIM and GLIDE methods, more aura-like structures are present. The remaining methods, DAGAN\_C, LOGAN, RSGAN-H, and Stable Diffusion have a rather similar fingerprint consisting of a single horizontal stripe. Finally, more or less concentrated energy peaks are present in almost all fingerprints. As expected, the fingerprint extracted from the pristine images lacks relevant geometric structures and is much less structured than those obtained from the fake ones. Taking inspiration from that, the proposed method injects credible but unspecific patterns in pristine images during models' training to make them learn without being tied to specific fingerprints.
\begin{table}[t]
    \centering
    \resizebox{\textwidth}{!}{%
    \begin{tabular}{l@{}@{\hspace{0.5cm}}lcccc|cc}
        \toprule
         & & \textbf{Grag. et al} \cite{uninagans} & \textbf{Grag. et al} \cite{uninagans} & \textbf{Wang et al} \cite{easyspot} & \textbf{Corvi et al} \cite{uninadms} & \textbf{Our (Resnet50)} & \textbf{Our (Swin-Base)} \\
        &  & StyleGAN2 & ProGAN & ProGAN & LDM & None & None \\
        \hline
        \hline
        \addlinespace
        \multirow{17}{*}{\rotatebox[origin=c]{90}{\textbf{Generative Adversarial Networks}}}& \textbf{BETA\_B} & 0.2 & 56.4 & 86.9 & 9.8 & 81.7 & \textbf{99.3} \\
        & \textbf{BICYCLEGAN} & 0.0 & 10.6 & 0.6 & 13.8 & \textbf{85.1} & 55.7 \\
        & \textbf{BIGGAN} & 9.3 & 44.2 & 10.3 & 24.9 & \textbf{60.6} & 53.3 \\
        & \textbf{DAGAN\_C} & 27.8 & 41.1 & 16.8 & 6.8 & 95.3 & \textbf{99.5} \\
        & \textbf{DFCVAE} & 0.6 & 8.2 & 34.8 & 0.1 & 81.4 & \textbf{100} \\
        & \textbf{DRIT} & 96.8 & \textbf{99.8} & 87.1 & 0.0 & 90.1 & 91.1 \\
        & \textbf{GANANIME} & 0.0 & 6.7 & 32.4 & \textbf{78.1} & 40.9 & 76.5 \\
        & \textbf{LOGAN} & 38.3 & 49.2 & 18.2 & 6.2 & \textbf{97.4} & \textbf{97.4} \\
        & \textbf{PIX2PIX} & 99.3 & 99.7 & 47.1 & 32.8 & \textbf{100} & 99.9 \\
        & \textbf{ProGAN} & 99.9 & \textbf{\underline{100}} & \textbf{\underline{100}} & 0.2 & 32.0 & 57.2 \\
        & \textbf{RelGAN} & \textbf{99.6} & 96.1 & 28.5 & 0.1 & 26.8 & 61.4 \\
        & \textbf{RSGAN-H} & 59.6 & 78.9 & 31.4 & 8.8 & \textbf{99.4} & 97.3 \\
        & \textbf{SAGAN} & 1.8 & \textbf{96.2} & 80.8 & 76.6 & 75.7 & 98.7 \\
        & \textbf{SRRNET} & 4.5 & 85.8 & 91.5 & \textbf{99.7} & 56.4 & 94.6 \\
        & \textbf{StyleGAN} & \textbf{100} & \textbf{100} & 99.3 & 1.3 & 31.6 & 61.9 \\
        & \textbf{StyleGAN2} & \textbf{\underline{100}} & 99.7 & 47.0 & 0.0 & 63.7 & 85.0 \\
        & \textbf{UNIT} & 90.2 & \textbf{91.3} & 19.5 & 13.6 & 77.5 & 55.6 \\
        \addlinespace
        \hline
        \addlinespace
        \multirow{8}{*}{\rotatebox[origin=c]{90}{\textbf{Diffusion Models}}} & \textbf{ADM} & 5.9 & 17.7 & 3.0 & 9.9 & 49.6 & \textbf{78.9} \\
        & \textbf{ADM-G} & 5.1 & 19.1 & 2.3 & 5.7 & 57.2 & \textbf{89.2} \\
        & \textbf{DDPM} & 51.1 & 99.7 & 17.0 & 0.1 & 27.5 & \textbf{71.2} \\
        & \textbf{DDIM} & 51.0 & \textbf{100} & 17.6 & 0.0 & 33.8 & 75.0 \\
        & \textbf{GLIDE} & 0.5 & 0.9 & 0.0 & \textbf{100} & 49.6 & 99.5 \\
        & \textbf{LDM} & 0.1 & 0.9 & 0.6 & \underline{35.2} & 39.5 & \textbf{88.6} \\
        & \textbf{Stable Diffusion} & 39.4 & 28.8 & 5.9 & 2.9 & 91.2 & \textbf{94.1} \\
        & \textbf{Stable Diffusion XL} & 0.1 & 0.1 & 1.5 & \textbf{99.8} & 36.3 & 96.3 \\
        \addlinespace
        \hline
        \addlinespace
        & \textbf{Pristine} & 99.9 & \textbf{100} & 99.3 & 99.9 & 95.3 & 97.4 \\
        \addlinespace
        \hline
        \hline
        \addlinespace
        \addlinespace
        & \textbf{Mean} & 40.9 & 58.9 & 37.7 & 27.9 & 64.4 & \textbf{83.6} \\
        \addlinespace
        \bottomrule
    \end{tabular}}
    \caption{Recall (for fake images) and specificity (for pristine images) obtained for each considered model on the test set. The first row under the model's names indicates the type of image generator used in the training set. The bold scores are the best obtained by the models on the specific method and the underlined ones are the cases when the generator is used in the training set of the model.}
    \label{tab:evaluation}
\end{table}
\subsection{Evaluation}
\subsubsection{Classification Performance}
\label{sec:classification}
The models trained with the proposed approach have been evaluated on the test set presented in Section \ref{sec:dataset} with other state-of-the-art deepfake detectors. 
In Table \ref{tab:evaluation}, we present the performance of each image generator, showing the recall metrics and specificity of pristine images.
The models we compare with are Resnet50 trained on datasets composed of pristine and fake images. Each literature's model used a different image generation approach for the construction of the training set's fake class, in particular, StyleGAN2 for \cite{uninagans}, ProGAN in \cite{uninadms} and Latent Diffusion for \cite{easyspot}. These models, therefore actually saw deepfakes during training, unlike ours, which are exclusively trained on pristine and pattern-injected images. As can be seen from the table, the proposed method yields excellent results on the test set. The trained Resnet50, and in particular the Swin-Base, are almost always able to recognize more than 70\% of the fake images in each subset while maintaining a specificity of 95.3\% and 97.4\%, respectively, on pristine images. The models we compared with, on the other hand, tend to have outstanding performances, mainly on the generators used to construct the fake class of the training set and on similar ones. This highlights how using fake images in the training set leads to an overfitting of the model on the used generator and to a lower generalization capacity in the detection of images synthesized in other ways. For example, the Resnet50 trained in \cite{uninagans} on pristine and StyleGAN2 images reaches around 100\% of recall when the generation approach remains the same or is a method which introduces similar fingerprints (shown in Figure \ref{fig:fingerprints_overview}) such as StyleGAN and ProGAN. Only occasionally it obtains good results on other methods and often fails to detect images generated with unseen generators.
An exception is made for the Corvi et al.\cite{uninadms} model which despite being trained on images generated via LDM, gets low recall on the same method. This behaviour can result from many factors such as different generator configurations, context and subjects of the generated images, number of iterations, etc.
Finally, looking at the overall mean calculated considering the results on both pristine and fake images obtained by each model the superiority of our approach is quite evident demonstrating strong generalization capabilities. In particular, the Swin-Base is capable of correctly classifying an average of 83.6\% images. 

\begin{figure}[t]
    \begin{minipage}[t]{0.48\linewidth}
        \centering
        \includegraphics[width=1\linewidth]{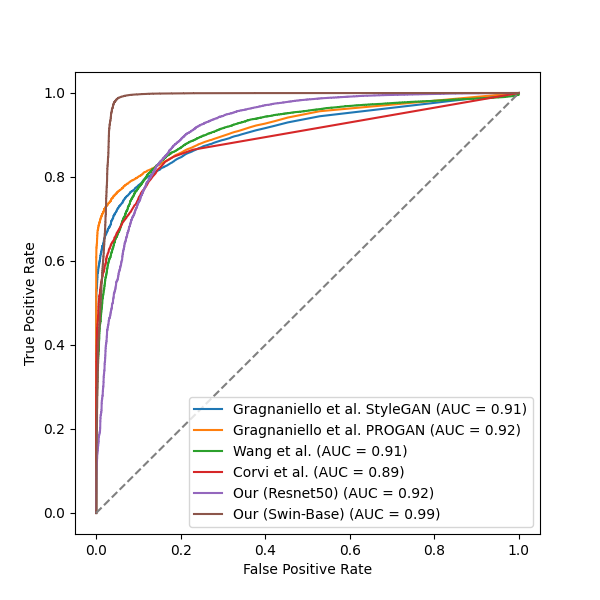}
        \caption{ROC Curve and AUC score of all considered models on the Test Set. }
        \label{fig:roc_curve}
    \end{minipage}\hfill 
    \begin{minipage}[t]{0.48\linewidth} 
        \centering
        
        \includegraphics[width=1\linewidth]{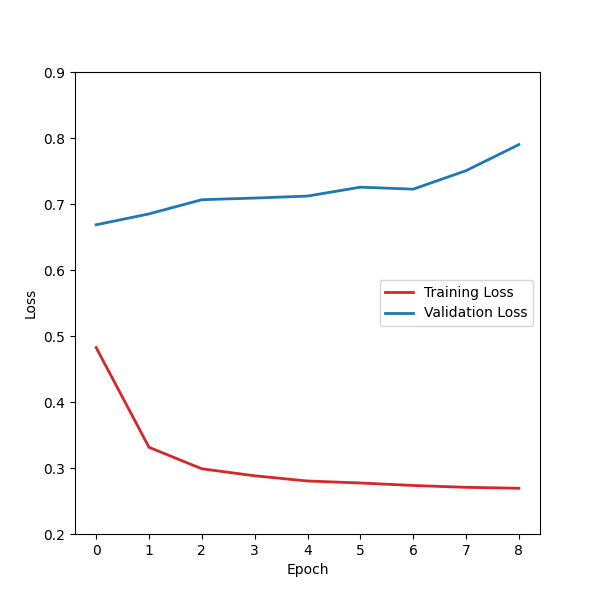}
        \caption{Training and Validation Loss for Swin-Base using the proposed approach.}
        \label{fig:loss}
    \end{minipage}
\end{figure}

To assess the effectiveness of our approach on the overall test set, in Figure \ref{fig:roc_curve} we show the ROC Curves and associated AUCs in comparison with the other considered techniques. The Resnet50 trained with our procedure, although considerably less specific on an image generation method, manages to maintain an AUC and ROC Curve rather similar to those trained with traditional methods, which tend to be more decisive in the classification of pristine images. The Swin-Base, on the other hand, has an AUC of 0.99 and a ROC Curve showing a greater ability to detect synthetic images than traditional methods.

In general, deepfake detectors are trained for many epochs by seeing the same images several times. In Figure \ref{fig:loss}, we show as an example a Swin-Base loss, trained with our approach, through epochs on the training set, composed of pristine and injected images, and on a validation set consisting of pristine and Stable Diffusion XL images. The model learns at the first epoch to distinguish between pristine and fake class, while it is deteriorating to train the model for many epochs. Continuing the training, the model tends to memorize the synthetic patterns, increasing the validation loss, and becoming unusable for deepfake detection. It is therefore necessary to perform a single epoch to obtain the best degree of generalization. Overcoming this limitation and training the models for many epochs without losing deepfake detection capabilities could lead to even better results than those already reported (e.g. reducing the learning rate or changing the model's architecture).

\begin{table}[t]
  \centering
    \resizebox{\textwidth}{!}{%
\def\arraystretch{1.5}
  \begin{tabular}{c|ccccccccccccccccc|cccccccc|c}
  & \multicolumn{17}{c}{Generative Adversarial Networks (GANs)} & \multicolumn{8}{c}{Diffusion Models (DMs)}  \\
    \hline
      & \rotatebox[origin=c]{90}{\textbf{BETA\_B}} & \rotatebox[origin=c]{90}{\textbf{BIGGAN}} & \rotatebox[origin=c]{90}{\textbf{BICYCLEGAN}} & \rotatebox[origin=c]{90}{\textbf{DAGAN\_C}} & \rotatebox[origin=c]{90}{\textbf{DFCVAE}} & \rotatebox[origin=c]{90}{\textbf{DRIT}} & \rotatebox[origin=c]{90}{\textbf{GAN\_ANIME}} & \rotatebox[origin=c]{90}{\textbf{LOGAN}} & \rotatebox[origin=c]{90}{\textbf{PIX2PIX}} & \rotatebox[origin=c]{90}{\textbf{ProGAN}} & \rotatebox[origin=c]{90}{\textbf{RelGAN}} & \rotatebox[origin=c]{90}{\textbf{RSGAN-H}} & \rotatebox[origin=c]{90}{\textbf{SAGAN}} & \rotatebox[origin=c]{90}{\textbf{SRRNET}} & \rotatebox[origin=c]{90}{\textbf{StyleGAN}} & \rotatebox[origin=c]{90}{\textbf{StyleGAN2}} & \rotatebox[origin=c]{90}{\textbf{UNIT}} & \rotatebox[origin=c]{90}{\textbf{ADM}} & \rotatebox[origin=c]{90}{\textbf{ADM-G}} & \rotatebox[origin=c]{90}{\textbf{DDPM}} & \rotatebox[origin=c]{90}{\textbf{DDIM}} & \rotatebox[origin=c]{90}{\textbf{GLIDE}} & \rotatebox[origin=c]{90}{\textbf{LDM}} & \rotatebox[origin=c]{90}{ \textbf{ Stable Diffusion } } & \rotatebox[origin=c]{90}{\textbf{Stable Diffusion XL}} & \rotatebox[origin=c]{90}{\textbf{Pristine}}
\\
      \hline
    \textbf{Mean} & 0.98 & 0.51 & 0.52 & 0.96 & 0.99 & 0.76 & 0.70 & 0.93 & 0.97 & 0.56 & 0.56 & 0.94 & 0.96 & 0.89 & 0.56 & 0.66 & 0.54 & 0.67 & 0.76 & 0.62 & 0.64 & 0.96 & 0.76 & 0.94 & 0.77 & 0.14 \\ 
    \hline
    \textbf{STD} & 0.07 & 0.14 & 0.10 & 0.08 & 0.00 & 0.17 & 0.23 & 0.13 & 0.06 & 0.17 & 0.14 & 0.14 & 0.10 & 0.16 & 0.16 & 0.15 & 0.18 & 0.21 & 0.19 & 0.19 & 0.19 & 0.08 & 0.18 & 0.14 & 0.14 & 0.12  \\
    \hline
  \end{tabular}
  }\\[5pt]
  \caption{Mean and Standard Deviation of Swin-Base's confidences, trained with the proposed method on each generation method of the test set.}
  \label{tab:predictions}
\end{table}

\begin{figure}[t]
    \begin{subfigure}[t]{\linewidth}
        \centering
        \includegraphics[width=1\linewidth]{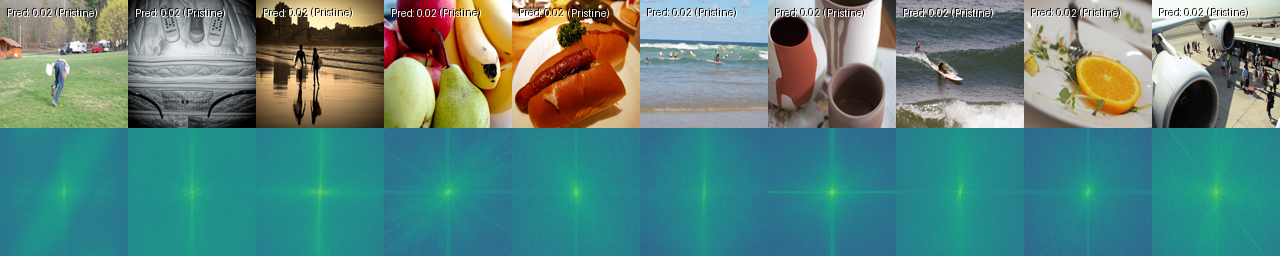}
        \caption{The most confident classifications on pristine images.}
        \label{fig:pristine_most_correct}
    \end{subfigure}\\[4pt] 
    \begin{subfigure}[t]{\linewidth} 
        \centering
        
        \includegraphics[width=1\linewidth]{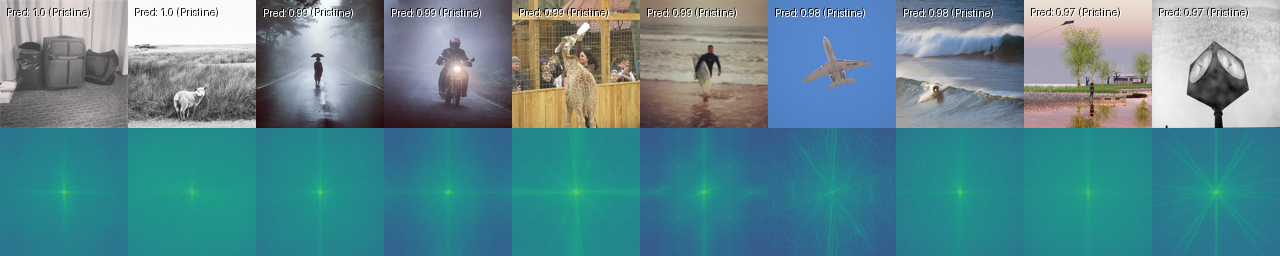}
        \caption{The most misclassified pristine images.}
        \label{fig:pristine_most_wrong}
    \end{subfigure}\\[4pt] 

    \begin{subfigure}[t]{\linewidth} 
        \centering
        
        \includegraphics[width=1\linewidth]{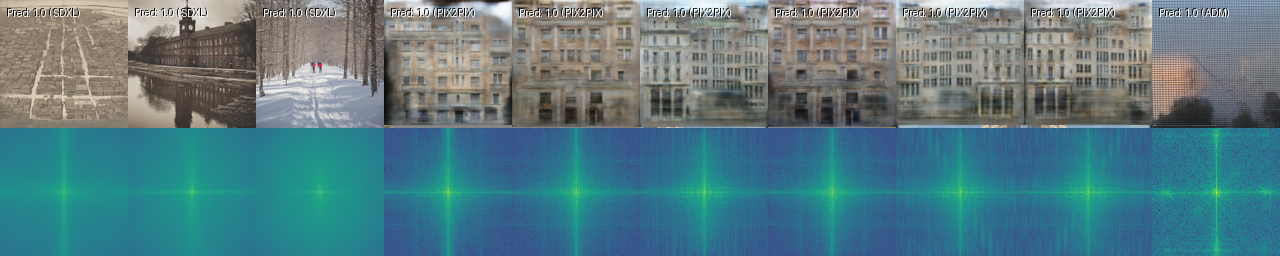}
        \caption{The most confident classifications on fake images.}
        \label{fig:fake_most_correct}
    \end{subfigure}\\[4pt] 

    \begin{subfigure}[t]{\linewidth} 
        \centering
        
        \includegraphics[width=1\linewidth]{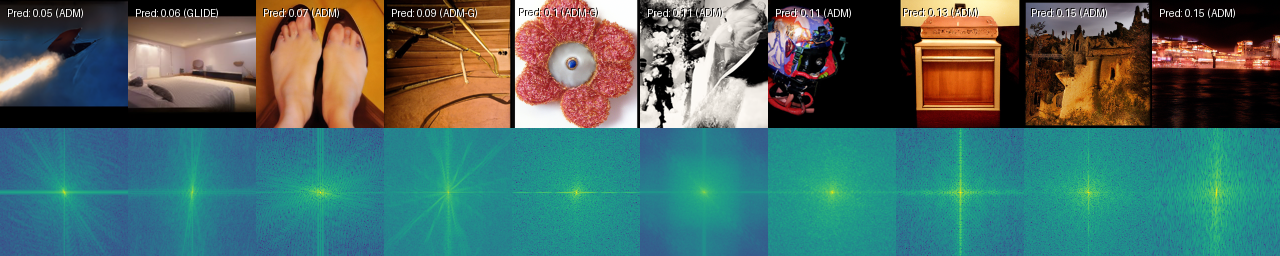}
        \caption{The most misclassified fake images.}
        \label{fig:fake_most_wrong}
    \end{subfigure}\\[3pt] 
    \caption{Best and worst ten images classified for each class by the Swin-Base detector trained with the proposed method. For each image, the model's prediction and the generation method or if the image is pristine.}
\end{figure}

\subsubsection{Error Analysis}
To investigate the confidence in the trained models' classifications, in Table \ref{tab:predictions} we report the averages and standard deviation of predictions made by our Swin-Base on the test set. It can be seen that although the model has never seen deepfake images in the training phase, it has sufficiently understood the concept of structured fingerprints, observing the injected synthetic patterns, and it is extremely confident in the prediction of many fake images. Examples in this regard are BETA\_B, DAGAN\_C, and DFCVAE, where the model's prediction is very near to 1. In some cases, on the other hand, the choice of threshold is crucial because on images generated with methods such as BIGGAN or BicycleGAN, the prediction of the model is very close to 0.5, and therefore, different thresholds could produce more or less false negatives. The model's predictions on pristine images, on the other hand, are mainly values very close to 0 indicating an excellent capability in their discrimination.

Finally, we can derive some insights from observing the test set images classified with more or less confidence by the Swin-Base model examined.
Figure \ref{fig:pristine_most_correct} shows the ten pristine images on which the model gave a prediction closest to zero, classifying them with the highest confidence. In their Fourier Transforms these images show no particular patterns and are therefore easily recognized as pristine by the model. On the other hand, the images shown in Figure \ref{fig:pristine_most_wrong} are the pristine ones on which the model gave a prediction closer to one, thus drastically mistaking them as fakes. It is possible to see how in some cases, albeit extremely rare, even pristine images can contain artifacts in the frequency domain even if they tend to be less pronounced than those found in fake images. This can mislead the model and generate sporadic false positives. 
In Figure \ref{fig:fake_most_correct}, we show the fake images predicted with the highest confidence by the model. These tend to be those produced by PIX2PIX as they have a very pronounced grid frequency pattern and in this case further accentuated by the geometries of the figures represented in the spatial domain. Interesting is the case of Stable Diffusion XL which although it does not seem to have a marked frequency pattern, the model effectively manages to distinguish it from pristine images. Finally, in Figure \ref{fig:fake_most_wrong}, we show the ten fake images in the test set on which the model gave a prediction closer to zero and thus mistakenly classified as pristine. It can be seen that almost all of them are generated with ADM or ADM-G. Their fingerprints do not represent particularly recognizable geometric patterns but something more distributed (as also shown in Section \ref{sec:fingerprint_analysis}) and thus not sufficiently reproduced by the patterns we introduced in the training phase.
From all these insights we can see how with a careful refinement of the injected patterns it could be even possible to further improve the discriminative ability of the models.

\section{Conclusions}
In this paper, we addressed the generalization problem in the field of synthetic image detection by proposing a novel approach for deepfake detectors' training. Exploiting the observations of frequency fingerprints typically introduced by image generation methods, we trained models using only pristine images that randomly undergo the injection of synthetic frequency patterns (inspired by geometric, grids, or aura shapes). The proposed method allowed us to train deepfake detectors without using any fake images, untying them from specific generation techniques. Our models have shown considerable superiority in terms of generalization over a wide range of image-generation methods in comparison with other state-of-the-art techniques. In future work, we aim to conduct a thorough investigation 
to identify the most effective synthetic patterns that enhance detection capabilities, improving the injection approach (e.g., considering also the phase or injecting in different color spaces) and exploring the possibility of using synthetic patterns for deepfake detectors fine-tuning.



\ifdoubleblind
\else
\subsubsection{Acknowledgments}
This work was partially supported by the following projects:
Tuscany Health Ecosystem (THE) (CUP B83C22003930001) and 
SERICS \\ (PE00000014, MUR PNRR - NextGenerationEU),
AI4Media (EC H2020 - n. 951911)
and AI4Debunk (Horizon EU n. 101135757),
FOSTERER (Italian MUR PRIN 2022).
We acknowledge the CINECA award under the ISCRA initiative, for the availability of high-performance computing resources and support. 
\fi

%
%
\bibliographystyle{splncs04}
\bibliography{egbib}
\end{document}